\documentclass{article}
\usepackage{spconf,amsmath,graphicx}

\usepackage{times}  
\usepackage{helvet} 
\usepackage{courier}  
\usepackage[hyphens]{url}  
\usepackage{graphicx} 
\usepackage{graphicx}  

\usepackage[utf8]{inputenc} 

\usepackage{booktabs}       
\usepackage{amsfonts}       
\usepackage{nicefrac}       
\usepackage{microtype}      

\usepackage{subfigure}
\usepackage{amsmath}
\usepackage{amsthm}

\usepackage{mathrsfs}

\usepackage{multirow}
\usepackage{paralist}

\usepackage{float}
\usepackage{booktabs}
\usepackage{multirow} 
\usepackage{amssymb}
\usepackage{graphicx}
\usepackage{subfigure} 
\usepackage{caption} 
\usepackage[linesnumbered,ruled,vlined]{algorithm2e}
\SetKwInput{KwInput}{Input}                
\SetKwInput{KwOutput}{Output}              
\usepackage[colorlinks,linkcolor=red,anchorcolor=blue,citecolor=blue,pagebackref=true]{hyperref}

\title{SAHDL: Sparse Attention Hypergraph Regularized Dictionary Learning}
\name{Shuai Shao, Rui Xu, Yan-Jiang Wang, Weifeng Liu, Bao-Di Liu
\sthanks{Corresponding Author. Thanks to the Natural Science Foundation of Shandong Province, China (Grant No. ZR2019MF073) for funding.}
}
\address{
College of Control Science and Engineering, China University of Petroleum (East China)\\
}

\begin{document}
%
\maketitle
\begin{abstract}
In recent years, the attention mechanism contributes significantly to hypergraph based neural networks. However, these methods update the attention weights with the network propagating. That is to say, this type of attention mechanism is only suitable for deep learning-based methods while not applicable to the traditional machine learning approaches. In this paper, we propose a hypergraph based sparse attention mechanism to tackle this issue and embed it into dictionary learning. More specifically, we first construct a sparse attention hypergraph, asset attention weights to samples by employing the $\ell_1$-norm sparse regularization to mine the high-order relationship among sample features. Then, we introduce the hypergraph Laplacian operator to preserve the local structure for subspace transformation in dictionary learning. Besides, we incorporate the discriminative information into the hypergraph as the guidance to aggregate samples. Unlike previous works, our method updates attention weights independently, does not rely on the deep network. We demonstrate the efficacy of our approach on four benchmark datasets.
\end{abstract}
In recent years, dictionary learning-based visual classification tasks have reached or even surpassed human beings' level. The ultimate goal of dictionary learning is to obtain an overcomplete dictionary to represent samples. Generally, pioneer dictionary learning algorithms, as examples LLC~\cite{wang2010locality}, SHDL~\cite{jenatton2010proximal}, introduce sparse codes to minimize the residual error to generalize dictionary bases (similar to cluster centers in k-means). However, the subspace learning is for each sample. It does not take notice of the relationships among different classes. In the following, discriminative information is introduced to solve this problem. LC-KSVD~\cite{jiang2013label}, FDDL~\cite{yang2011fisher}, RFDDL~\cite{zhang2019joint}, LEDL~\cite{shao2020label}~\emph{et al.} incorporated discriminative information as the constraint term into the objective function. In addition, CSDL~\cite{liu2016face}, CDLF~\cite{wang2020class}, TP-DPL~\cite{zhang2019learning}~\emph{et al.} implicitly employed the discriminative information by building class-specific dictionary. Each method tries to increase the distance among classes for classification. However, all approaches mentioned above do not consider the distribution of data. As one kind of subspace learning algorithms, it is difficult to keep the distribution after subspace transformation consistent with the original data. 

To tackle this problem, Zheng~\emph{et al.} proposed GraphSC \cite{zheng2010graph} to embed the graph based manifold regularization into the objective function of dictionary learning. The Graph is a powerful way to represent pair-wise relations for samples via the adjacency matrix. Therefore, the graph based manifold structure is beneficial to maintain the local structure of the original data. While in the real world, high-order relationships among different samples (for example, one type of relationship between two persons according to their age and another type of relationship according to their gender) can not be modeled by a simple graph. Therefore, Zhou~\emph{et al.} proposed the hypergraph structure~\cite{zhou2007learning} to represent the data correlation accurately. Hypergraph consists of a vertex set (e.g., samples) and a hyperedge set. Each hyperedge includes a flexible number of vertices. We generally use an incidence matrix to formulate the hypergrah. The structure is capable of modeling the high-order relationship mentioned above. Notably, a hypergraph is the same as a simple graph when each hyperedge's degree is restricted to $2$. Following, Gao~\emph{et al.} proposed the HLSC~\cite{gao2012laplacian} to introduce hypergraph structure to manifold regularization, which is beneficial to preserve high-order information intact. 

However, although the HLSC method has been widely developed in various tasks, there are still two significant challenges. The first one is a general challenge for all hypergraph based approaches: samples in a hyperedge playing equally important roles for the central one in image classification tasks. Most classical methods, such as HLSC, CDMH~\cite{zhang2018cross}, and HGNN~\cite{feng2019hypergraph}, construct the hypergraph according to the similarity of sample features via knn. All the samples have the same weights among the hypergraph. While in many cases, samples with similar features belong to different categories (as an example, tigers and cats), utilizing the way mentioned above may result in faulty connections. One considerable solution is to provide different attention weights (for example, assigning tigers low weights when classifying cats) for samples. Besides, several research studies, including FSL-GNN~\cite{VictorFew-shot}, HLPN~\cite{zhang2020hypergraph}, and DHCF~\cite{ji2020dual} designed novel mechanisms for this problem, but the attention values in these approaches are updated with the network propagating. These approaches are only suitable for deep architecture, not applicable to machine learning tasks. 

The second challenge is to embed the discriminative information into dictionary atoms appropriately in supervised and semi-supervised learning tasks. 
Compared with several traditional dictionary learning methods such as LC-KSVD, LEDL, there is no proper method to embed the discriminative information for hypergraph manifold based dictionary learning.

To address these two challenges, we propose a novel algorithm named Sparse Attention Hypergraph regularized Dictionary Learning (SAHDL). On the one hand, we present a sparse attention mechanism to mark the crucial samples for building a hypergraph. It is helpful to mine the samples' essential attributes and then treat them as the weights to the corresponding connection. The final score is determined by the point product of samples' similarity and weight. We describe this modal as a sparse attention feature (saf) modal. 

On the other hand, as hypergraph structure can fuse multi-modal information, we incorporate discriminative information as another modal, named label (lb) modal. Specifically, each hyperedge denotes a category, contains all the labeled nodes that belong to this class. It is also efficient to introduce discriminative information by embedding the one hot label matrix into the objective function, just like mHDSC~\cite{liu2014multiview}. However, compared with the proposed SAHDL approach, mHDSC presents additional parameters (need to be manually adjusted) and variables (need to be optimized), consume more resources. 

\begin{figure*}
	\begin{center}
		\includegraphics[width=0.8\linewidth]{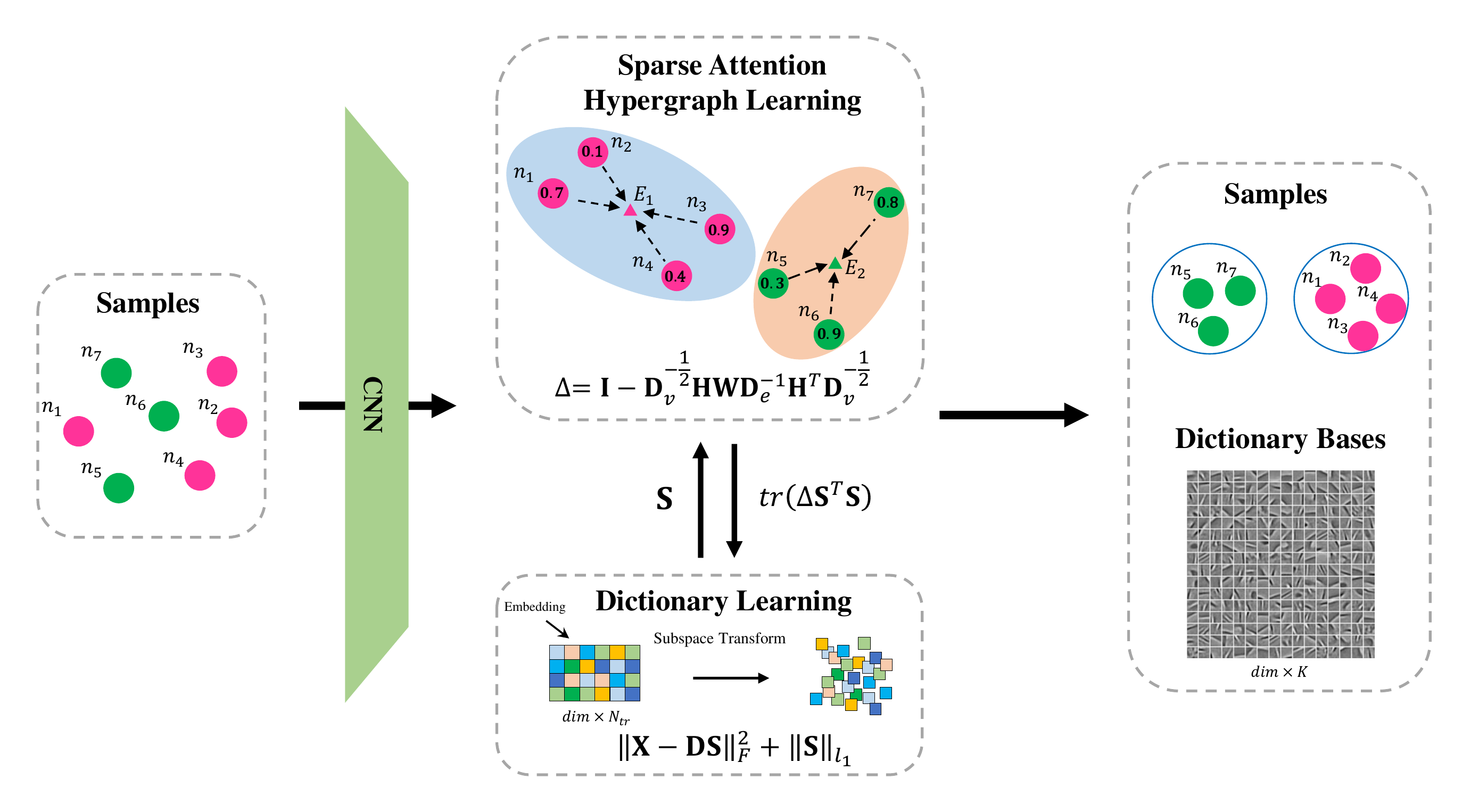}
	\end{center}
	\caption{The SAHDL framework. It contains two modules: $i)$ Sparse attention hypergraph learning module, focus on capturing the relationships among samples and aggregating hyperedge information by Laplacian regularization. $E_1$ and $E_2$ denote the aggregated vertices. Different backgrounds mean different hyperedges. $ii)$ Dictionary learning module, transforms the feature embeddings into a subspace. They propagate information by Equation~(\ref{equation: Objective_function_inductive}). After that, we obtain dictionary bases to represent samples, which is helpful for classification. More details please refer to section~\ref{Section: Methodology}.}
	\label{figure: SAHDL}
\end{figure*}

In the end, we introduce a hypergraph Laplacian operator to embed the designed hypergraph manifold to constrain the subspace transformation in dictionary learning. Transductive and inductive based dictionary learning are both described in this paper. Figure~\ref{figure: SAHDL} illustrates the design of SAHDL.

In summary, the main contributions focus on:

\begin{itemize}
\item 
We make it possible to mine essential attributes of image features for constructing a hypergraph. We design a sparse attention mechanism to evaluate the importance of each vertex. Then we capture the corresponding attention weight. Thus, the manifold structure can aggregate information more efficiently, especially remarkable on the datasets without inherent manifold-structures and high-quality features. 
\item The proposed attention block is a model-agnostic structure. In this paper, we introduce it to hypergraph based dictionary learning. It is also suitable for graph/hypergraph based deep architecture, such as GCN~\cite{kipf2017semi}, HGNN~\cite{feng2019hypergraph}.

\item This dictionary learning model incorporates the discriminative information into hypergraph manifold. This way achieves ideal results while reducing the cost of calculation.

\item Experimental results demonstrate that the proposed sparse attention hypergraph regularized dictionary learning method significantly improves the classification performance compared with other stat-of-the-art dictionary learning methods.
\end{itemize}

\section{Related Work}
\label{Section: Related work}
In this section, we first introduce the attention mechanism on graph based works, then review the manifold structure regularized dictionary learning.

\subsection{Attention on Graph based Works}
\label{Section: Attention mechanism}
The attention mechanism is widely used in various areas in recent years, capable of focusing on a certain region. The attention mechanisms are helpful in graph based works and can be split into two categories: $i)$ Chanel based attention, such as GAT~\cite{velivckovic2018graph}, KGAT~\cite{wang2019kgat}, HGAN~\cite{wang2019heterogeneous}, works on the neurons of a deep network, update the edge weights of samples by aggregation and propagation of sample information. $ii)$ Sample-based attention. It employs the attributes of samples to construct attention mechanism, including Kernel-GAT~\cite{liu2020fine}, HAN~\cite{kim2020hypergraph}, ReGAT~\cite{li2019relation}, FSL-GNN~\cite{VictorFew-shot}, HLPN~\cite{zhang2020hypergraph}, DHCF~\cite{ji2020dual}. This category of approaches introduces a priori knowledge, which can help the graph structure better express the sample's characteristics to complete different tasks. The proposed SAHDL approach is part of sample-based attention. Unlike the above methods, the proposed sparse attention mechanism can update the weight without relying on the network propagating, suitable for more extensive fields such as traditional machine learning.

\subsection{Manifold Structure Regularized Dictionary Learning}
\label{Section: Manifold Structure on Dictionary Learning}
It is a general way to combine manifold learning with dictionary learning to preserve the local structure unchanged after subspace transformation. We can choose a graph or a hypergraph to represent the local structure. The Graph is capable of representing pair-wise relations, and the edge links two samples. However, the hypergraph can mine high-order relationships, and a hyperedge is involved in multiple samples. Besides, manifold regularization, such as LLE, Laplacian, p-Laplacian, Hessian, is introduced as a link among different subspaces. Many classical related works, containing \cite{zheng2010graph,liu2013learning,liu2014multiview,ma2018hypergraph}, have demonstrate that different regularization methods are suitable for different datasets. In this paper, we adopt the Laplacian operator for preserving the local structure.

\section{Methodology}
\label{Section: Methodology}

This section focuses on introducing the Sparse Attention Hypergraph Regularized Dictionary Learning (SAHDL) algorithm. To be specific, first, we briefly review the original dictionary learning method. Second, we construct the sparse attention hypergraph in detail. Third, the Laplacian operator is adopted to preserve the hypergraph manifold. Fourth, the objective function of the SAHDL algorithm, as well as the optimization strategy, are designed. Finally, we discuss and conclude the SAHDL algorithm.

\subsection{Review of Dictionary Learning Method}
\label{Section: Dictionary_Learning}
We assume that the training data be $\mathbf{X}=[{\mathbf{x}}_1,{\mathbf{x}}_2,\dots,{\mathbf{x}}_{N_{tr}}] \in \mathbb{R}^{dim \times N_{tr}}$ and the testing data be $\mathbf{Y} =[{\mathbf{y}}_1,{\mathbf{y}}_2,\dots,{\mathbf{y}}_{N_{ts}}] \in \mathbb{R}^{dim \times N_{ts}}$, where ${\mathbf{x}}_i$, ${\mathbf{y}}_i$ ($i = 1, 2, \dots$) is the feature embedding of the $i_{th}$ sample, and the $dim$ denotes the dimension of each sample. $N_{tr}$ and $N_{ts}$ represent the number of training samples and testing samples, respectively.

In contrast to conventional matrix factorization methods, dictionary learning method aims to find a sparse representation $\mathbf{S}=[{\mathbf{s}}_1,{\mathbf{s}}_2,\dots,{\mathbf{s}}_{N_{tr}}] \in{\mathbb{R}}^{K\times {N_{tr}}}$ while updating an over-complete dictionary $\mathbf{D}=[{\mathbf{d}}_1,{\mathbf{d}}_2,\dots,{\mathbf{d}}_K] \in{\mathbb{R}}^{dim\times K}$. Here, $K$ is the number of atoms in dictionary. $\mathbf{s}_i$ ($i = 1, 2, \dots$) represents the sparse representation of the $i_{th}$ sample in $\mathbf{X}$. $\mathbf{d}_k$ represents the $k_{th}$ atoms in dictionary $\mathbf{D}$. The objective function of dictionary learning method can be formulated as follows,

\begin{equation}
\begin{split}
         \mathop {\arg \min}\limits_{{\mathbf{D}},{\mathbf{S}}} f_1({\mathbf{D}},{\mathbf{S}}) =
        \left\| \mathbf{X} - \mathbf{D}\mathbf{S} \right\|_F^2 + 2\alpha \left\| \mathbf{S} \right\|_{\ell_1}\\
		{\rm{s}}.t.{\kern 4pt}\left\| {\mathbf{d}}_{k} \right\|_2^2 \le 1, \left( {k = 1,2, \cdots K} \right)
\end{split}
\label{equation: DL_traditional}
\end{equation}
where $\left\| \mathbf{X} - \mathbf{D}\mathbf{S} \right\|_F^2$ is the reconstruction error term and $\| \bullet \|_F$ represents the Frobenius norm. $\left\| \mathbf{S} \right\|_{\ell_1}$ represents the sparse constraint for $\mathbf{S}$ (i.e., $\sum_{i=1}^{N_{tr}}\|\mathbf{s}_i\|_1$). $\alpha$ is a trade-off between the reconstruction error and the sparsity.



\subsection{Sparse Attention Hypergraph Construction}
\label{Section: Hypergraph_Construction}
How to construct a suitable graph structure is the most critical procedure for any graph based tasks. Unlike the regular graph, the hypergraph can capture high-order relationships by collecting similar samples into the same hyperedge. Figure~\ref{figure: G_HG_SAHG} shows the comparison between the regular graph and the hypergraph. We define the hypergraph as $\mathcal{G}=(\mathcal{V},\mathcal{E},\mathbf{W})$. Here, $\mathcal{V}$ denotes the vertex set, and each vertex represents a sample. $\mathcal{E}$ is the hyperedge set, and $\mathbf{W}$ is a diagonal matrix where each element in $\mathbf{W}$ represents the weight of the corresponding hyperedge. In this paper, we give equal weights for all the hyperedges.
The connection of hyperedges and vertices can be represented by the incidence matrix $\mathbf{H}\in{\mathbb{R}}^{|\mathcal{V}|\times{|\mathcal{E}|}}$. Besides, the hypergraph can fuse multi-modal information by concatenating multiple views' hyperedges.

In this paper, we utilize two modal information, including the sparse attention feature (saf) modal and the label (lb) modal, to construct the hypergraph. Thus, the hyperedge $\mathcal{E}$ and the incidence matrix $\mathbf{H}$ are defined as $\mathcal{E}=[\mathcal{E}_{saf},\mathcal{E}_{lb}]$ and $\mathbf{H}=[\mathbf{H}_{saf},\mathbf{H}_{lb}]$, where $\mathbf{H}_{saf}\in{\mathbb{R}}^{|\mathcal{V}|\times{|\mathcal{E}_{saf}|}}$, $\mathbf{H}_{lb}\in{\mathbb{R}}^{|\mathcal{V}|\times{|\mathcal{E}_{lb}|}}$. 

\subsubsection{\textbf{Sparse Attention Feature Modal}}
\label{Section: sparse_attention_feature_modal}
In the sparse attention feature modal, we construct the connections by mining the high-order relationships among samples. Specifically, we put the vertices with similar features into the same hyperedge via the k-nearest-neighbor (knn) approach. The final scores in the incidence matrix are defined as follows,

\begin{equation}
\begin{split}
        \mathbf{H}_{saf}=
        \left\{\begin{array}{cc}
            {\exp \left(-dis\left({v_i}, {v}_{c}\right)^{2}\right)} \times z_i & {\text { if } {v_i} \in {e}_{saf}} \\
            {0} & {\text{ o.w. }}
        \end{array}\right.
\end{split}
\label{equation: elements_in_H_{saf}}
\end{equation}
where ${e}_{saf}$ is a hyperedge in $\mathcal{E}_{saf}$, ${v_i}$ denotes the $i_{th}$ vertex in $\mathcal{V}$ and ${v}_c$ is the central vertex in ${e}_{saf}$. $dis$ denotes the operator to compute the distance. $z_i$ represents the attention weight for the $i_{th}$ vertex. We compute $z_i$ via the following formula,

\begin{equation}
\begin{split}
         \mathop {\arg \min}\limits_{\mathbf{ z}} f_2(\mathbf{ z}) =
        \left\| \mathbf{x} - \mathbf{P}\mathbf{ z} \right\|_F^2 + 2\epsilon \left\| \mathbf{ z} \right\|_{\ell_1}\\
\end{split}
\label{equation: sparse_attention}
\end{equation}
where $\mathbf{ z}=[ z_1, z_2, \dots,  z_{N_{sa}}] \in \mathbb{R}^{N_{sa} \times 1}$ is the weight vector. The following section~\ref{section: Optimization for Sparse Attention Module} provides the detailed optimization strategy. $\mathbf{P} = knn(\mathcal{V}) \in \mathbb{R}^ {dim \times N_{sa}}$, denotes the feature embedding of samples selected from the k-nearest-neighbors, $N_{sa}$ is the number of nearest neighbors. $\epsilon$ is the constant parameter to control the sparsity. 

\subsubsection{\textbf{Label Modal}}
\label{Section: label_modal}

For the label modal hypergraph construction, we aggregate the vertices with the same labels into one hyperedge. Notably, introducing discriminative information in this form has the equal effect of directly adding a label constraint term to objective function such as mHDSC, LEDL (more validation is shown in the experimental section~\ref{section: ablation study}). The elements in the incidence matrix are defined as follows:

\begin{equation}
\begin{split}
        \mathbf{H}_{lb}=
        \left\{\begin{array}{cc}
            {1} & {\text { if } {v} \in {e}_{lb}} \\
            {0} & {\text{ o.w. }}
        \end{array}\right.
\end{split}
\label{equation: elements_in_H_{lb}}
\end{equation}
where ${e}_{lb}$ denotes a hyperedge in $\mathcal{E}_{lb}$. Note that, in semi-supervised tasks, the elements in $\mathbf{H}_{lb}$ are set to $1$ when the corresponding vertices have labels. Otherwise, they are set to $0$. 

\subsection{Hypergraph Manifold Regularization}
\label{Section: hypergraph_manifold_regularization}

Before introducing the Laplacian operator, we define two diagonal matrices as $\mathbf{D}_v$ (vertex degree matrix) and $\mathbf{D}_e$ (hyperedge degree matrix). The vertex degree $\delta(\mathbf{e})$ and the hyperedge degree $d(\mathbf{v})$ are formulated as:

\begin{equation}
\begin{split}
        \left\{\begin{array}{ll}
            \delta({e})= \sum_{{v} \in \mathcal{V}} \mathbf{H}({v},{e}) \\
            d({v})= \sum_{{e} \in \mathcal{E}} \mathbf{W}({e} ) \mathbf{H}({v},{e})
        \end{array}\right.
\end{split}
\label{equation: vertex_hyperedge_degree}
\end{equation}

To preserve the hypergraph based local structure for dictionary learning, we introduce the hypergraph regularizer and formulate the relationship between $\mathbf{H}$ and $\mathbf{S}$ as follows:

\begin{equation}
    \begin{split}
    &f_3\left(\mathbf{S}\right)\\
    &=\frac{1}{2}\sum_{c=1}^{C} \sum_{e \in \mathcal{E}} \sum_{u,v \in \mathcal{V}} 
    \frac{\mathbf{W}(e)\mathbf{H}(u,e)\mathbf{H}(v,e)}{\delta(e)}
    \left( \frac{\mathbf{S}\left(u,c\right)}{\sqrt{d\left( u \right)}}
    - \frac{\mathbf{S}\left(v,c\right)}{\sqrt{d\left( v \right)}} \right)^2\\
    &=\sum_{c=1}^{C} \sum_{e \in \mathcal{E}} \sum_{u,v \in \mathcal{V}}
    \frac{\mathbf{W}(e)\mathbf{H}(u,e)\mathbf{H}(v,e)}{\delta(e)}
    \left( \frac{\mathbf{S}\left(u,c\right)^2}{d\left( u \right)}
    - \frac{\mathbf{S}\left(u,c\right)\mathbf{S}\left(v,c\right)}{\sqrt{d\left( u \right)d\left( v \right)}} \right)\\
    &=\sum_{c=1}^{C} \sum_{u \in \mathcal{V}} \mathbf{S}\left(u,c\right)^2 
    \sum_{e \in \mathcal{E}}
    \frac{\mathbf{W}(e)\mathbf{H}(u,e)}{d(u)}
    \sum_{v \in \mathcal{V}} \frac{\mathbf{H}(v,e)}{\delta(e)}\\
    &{\kern 10pt}-\sum_{e \in \mathcal{E}}\sum_{u,v \in \mathcal{V}}
    \left( 
    \frac{\mathbf{S}\left(u,c\right)
    \mathbf{H}(u,e)
    \mathbf{W}(e)
    \mathbf{H}(v,e)
    \mathbf{S}\left(v,c\right)}
    {\sqrt{d\left( u \right)d\left( v \right)} \delta(e)
    } \right )\\
    &=\text{tr}\left(
    \left(
    \mathbf{I} - \mathbf{D}_v^{-\frac{1}{2}} \mathbf{H} \mathbf{W} \mathbf{D}_e^{-1} \mathbf{H}^T \mathbf{D}_v^{-\frac{1}{2}}
    \right)
    \mathbf{S}^T \mathbf{S}
    \right)\\
    &=\text{tr}\left(  \mathbf{\Delta} \mathbf{S}^T \mathbf{S}
    \right)
    \end{split}
\label{equation: laplace_regularization}
\end{equation}
where $\mathbf{\Delta}=\mathbf{I} - \mathbf{D}_v^{-\frac{1}{2}} \mathbf{H} \mathbf{W} \mathbf{D}_e^{-1} \mathbf{H}^T \mathbf{D}_v^{-\frac{1}{2}}$ denotes the normalized hypergraph Laplacian operator. 

\subsection{Sparse Attention Hypergraph Regularized Dictionary Learning (SAHDL)}
\label{Section: SAHDL}
\subsubsection{\textbf{Objective Function}}
We embed the hypergraph regularization term into the conventional dictionary learning method and form the objective function of SAHDL as follows,

\begin{equation}
\begin{split}
        &\mathop {\arg \min}\limits_{\mathbf{D},\mathbf{S}}     
        \mathcal{F}(\mathbf{D},\mathbf{S})
        = f_1(\mathbf{D},\mathbf{S}) + \beta {\kern 2pt} f_3(\mathbf{S})\\
		&=\left\| \mathbf{X} - \mathbf{D}\mathbf{S} \right\|_F^2 
		+ 2\alpha \left\| \mathbf{S} \right\|_{\ell_1} 
		+ \beta {\kern 2pt} \text{tr} \left( \mathbf{\Delta} \mathbf{S}^T \mathbf{S} \right)\\
		&{\rm{s}}.t.\left\| {{{\mathbf{d}}_{  k}}} \right\|_2^2 \le 1 {\kern 4pt} \left( {k = 1,2, \cdots K} \right)\\
\end{split}
\label{equation: Objective_function_inductive}
\end{equation}
where $\beta$ is to balance between the reconstruction error and the hypergraph regularization. We can obtain the optimal $\mathbf{D}$ and $\mathbf{S}$ via an alternative optimization strategy. The following section~\ref{section: Optimization} shows the solution details.

\subsubsection{\textbf{Classification}}
\label{section: Classification}
In the testing stage, one way for classification is to design a classifier plane $\mathbf{B}$ by using the optimal $\mathbf{S}$ to fit the one hot label matrix. 

First, we obtain the representation $\mathbf{S}_{ts}$ of the testing samples $\mathbf{Y}$ with the fixed dictionary $\mathbf{D}$, which can be formulated as,

\begin{equation}
\begin{split}
        \mathop {\arg \min}\limits_{\mathbf{S}_{ts}}
		\left\| \mathbf{Y} - \mathbf{D}\mathbf{S}_{ts} \right\|_F^2 
		+ 2\gamma \left\| \mathbf{S}_{ts} \right\|_{\ell_1} 
\end{split}
\label{equation: Classification}
\end{equation}
where $\mathbf{S}_{ts} \in \mathbb{R}^{K \times N_{ts}}$ denotes the learned sparse embeddings for the testing data.

Second, we predict the label of the testing samples $\mathbf{Y}$ via calculating the minimum value of the following formula,

\begin{equation}
\begin{split}
\mathbf{O} = \mathbf{B} \mathbf{S}_{ts}
\end{split}
\label{equation: predict}
\end{equation}

\subsection{Optimization}
\label{section: Optimization}
\subsubsection{\textbf{Optimization for Sparse Attention Hypergraph}}
\label{section: Optimization for Sparse Attention Module}

Observing Equation~\ref{equation: sparse_attention}, $\mathbf{P}$ is a constant feature embedding matrix for each sample. We introduce the Alternating Direction Method of Multipliers (ADMM)~\cite{boyd2011distributed} framework to solve the optimization problem.

\textbf{$i)$} Introducing an auxiliary variable ${\bf{q}}$ and reformulate Equation~(\ref{equation: sparse_attention}) as:
\begin{equation}
\begin{split}
        \mathop {\arg \min}\limits_{\mathbf{z},\mathbf{q}} f_4(\mathbf{z},\mathbf{q}) =
        \left\| \mathbf{x} - \mathbf{P}\mathbf{z} \right\|_F^2 + 2\epsilon \left\| \mathbf{q} \right\|_{\ell_1}
        {\kern 5pt} s.t. {\kern 2pt} \mathbf{z} = \mathbf{q}
\end{split}
\label{equation: ADMM1}
\end{equation}

\textbf{$ii)$} Writing the Lagrangian function of Equation~(\ref{equation: ADMM1}) as:
\begin{equation}
\begin{split}
        &\mathop {\arg \min}\limits_{\mathbf{z},\mathbf{q},\mathbf{m}} f_5(\mathbf{z},\mathbf{q},\mathbf{m}) \\
        &= \left\| \mathbf{x} - \mathbf{P}\mathbf{z} \right\|_F^2 
        + 2\epsilon \left\| \mathbf{q} \right\|_{\ell_1}
        + \left<\mathbf{m},  \mathbf{z} - \mathbf{q} \right> 
        + \rho \left\| \mathbf{z} - \mathbf{q} \right\|_F^2
\end{split}
\label{equation: ADMM2}
\end{equation}
where $\mathbf{m}$ is the augmented lagrangian multiplier, and $\rho>0$ denotes the penalty parameter. 

\textbf{$iii)$} Alternative updating $\mathbf{z}$, $\mathbf{q}$, $\mathbf{m}$ until Equation~(\ref{equation: ADMM2}) convergence, the solutions are formulated as follows,

\begin{equation}
\begin{split}
        \left\{\begin{array}{lll}
            \mathbf{z} 
            = \left( \mathbf{P}^T \mathbf{P} + \rho \mathbf{I} \right)^{-1}
            \left( \mathbf{P}^T \mathbf{x} + \rho {\kern 2pt} \mathbf{q} - \mathbf{m} \right)\\
            \mathbf{q} 
            = max \left\{ \mathbf{z} + \frac{\mathbf{m} - \epsilon\mathbf{I}}{\rho}, \mathbf{0} \right\}
            +min  \left\{ \mathbf{z} + \frac{\mathbf{m} - \epsilon\mathbf{I}}{\rho}, \mathbf{0} \right\}\\
            \mathbf{m} 
            = \mathbf{m} + \theta \left( \mathbf{z} - \mathbf{q} \right)  
        \end{array}\right.
\end{split}
\label{equation: ADMM_solutions}
\end{equation}
where $\mathbf{I}$ denotes the identity matrix, $\mathbf{0}$ is the zero vector, and $\theta$ denotes the gradient degree.

\subsubsection{\textbf{Optimization for the SAHDL}}
\label{section: Optimization for Objective Function}
As we can see, the ${\mathbf{D}}$ and ${\mathbf{S}}$ in the Equation (\ref{equation: Objective_function_inductive}) are not jointly convex. We employ the alternative optimization scheme to find the sub-optimal solutions for each variable with the other fixed to solve this problem. \\
\textbf{$i)$} Update $\mathbf{S}$ with $\mathbf{D}$ fixed, the sub-problem can be written as:

\begin{equation}
\begin{split}
        &\mathop {\arg \min}\limits_{\mathbf{S}} \mathcal{F}(\mathbf{S})\\
		&=\left\| \mathbf{X} - \mathbf{D}\mathbf{S} \right\|_F^2 
		+ 2\alpha \left\| \mathbf{S} \right\|_{\ell_1} 
		+ \beta \text{tr}\left( \mathbf{\Delta} \mathbf{S}^T \mathbf{S}\right)\\
		&=\sum_{n=1}^{N_{tr}}\mathbf{S}_{\bullet n}^T\mathbf{D}^T\mathbf{D}\sum_{n=1}^{N_{tr}}\mathbf{S}_{\bullet n}
		-2\sum_{n=1}^{N_{tr}}\left(\mathbf{X}^T\mathbf{D}\right)_{n \bullet}\mathbf{S}_{\bullet n}\\
		&{\kern 10pt}+ 2\alpha \sum_{k=1}^K\sum_{n=1}^{N_{tr}}|\mathbf{S}_{kn}|
		+\beta {\kern 2pt} \mathbf{\Delta}\sum_{k=1}^{K} \left(\mathbf{S}^T\mathbf{S}\right)_{k \bullet} \\  
\end{split}
\label{equation: optimization_UpdateS_Simplification}
\end{equation}
the closed-form solution of $\mathbf{S}$ is:
\begin{equation}
\begin{split}
        \mathbf{S}_{kn}=
            {\frac{max\left(\mathcal{J},\alpha \right) + min\left(\mathcal{J},\alpha \right)}
            {\left(\mathbf{D}^T\mathbf{D}
            \right)_{kk}
            +\beta\left(\mathbf{\Delta}\right)_{nn}}}
\end{split}
\label{equation: optimization_UpdateS_Skn_1}
\end{equation}
where 
\begin{equation}
\begin{split}
   \mathcal{J}&=\left(\mathbf{D}^T\mathbf{X} \right)_{kn}
    -\beta \sum_{r=1,r\neq n}^{N_{tr}} \left(\mathbf{\Delta}\right)_{nr}\mathbf{S}_{kr}
    -\sum_{l=1,l\neq k}^{K} \left(\mathbf{D}^T\mathbf{D}
    \right)_{kl} \mathbf{S}_{ln}
\end{split}
\label{equation: optimization_UpdateS_J}
\end{equation}
\textbf{$ii)$} Update $\mathbf{D}$ with $\mathbf{S}$ fixed, we rewrite the sub-problem as:

\begin{equation}
\begin{split}
        &\mathop {\arg \min}\limits_{\mathbf{D}} \mathcal{F}(\mathbf{D})
		=\left\| \mathbf{X} - \mathbf{D}\mathbf{S} \right\|_F^2 \\
		&{\kern 5pt}{\rm{s}}.t.\left\| {{{\bf{d}}_{ k}}} \right\|_2^2 \le 1
		 {\kern 4pt} \left( {k = 1,2, \cdots K} \right)\\
\end{split}
\label{equation: optimization_UpdateD_Simplification}
\end{equation}
According to the blockwise coordinate descent (BCD) algorithm \cite{liu2014blockwise}, we can directly obtain $\mathbf{D}$ as follows:
\begin{equation}
\begin{split}
        \mathbf{D}_{\bullet k} 
        = \frac{\mathbf{X}\left(\mathbf{S}_{k \bullet} \right)^T
        -\mathbf{\tilde{D}}^k \mathbf{S} \left(\mathbf{S}_{k \bullet} \right)^T}
        {\| \mathbf{X}\left(\mathbf{S}_{k \bullet} \right)^T
        -\mathbf{\tilde{D}}^k \mathbf{S} \left(\mathbf{S}_{k \bullet} \right)^T \|_2}
\end{split}
\label{equation: optimization_UpdateD}
\end{equation}
where $ \mathbf{\tilde{D}}=
        \left\{ \begin{array}{cc}
             {\mathbf{D}_{\bullet m}}  & {m \neq{k}}  \\
             {\mathbf{0}} & {m = k}
         \end{array}\right.$. We summarize the overall flowchart of SAHDL in Algorithm~\ref{Algorithm: SAHDL}.

\begin{algorithm}[t]
\DontPrintSemicolon
  
  \KwInput{$\mathbf{X} \in \mathbb{R}^{dim \times N_{tr}}$}
  \KwOutput{$\mathbf{D} \in \mathbb{R}^{dim \times K}$, $\mathbf{S} \in \mathbb{R}^{K \times N_{tr}}$}
    Construct saf-hypergraph $\mathbf{H}_{saf}$ by \textbf{Equation (\ref{equation: elements_in_H_{saf}})}.\\

    \While{i < maxitem}
    {
        Employ ADMM to optimize attention weights $\mathbf{z}$ by \textbf{Equation (\ref{equation: ADMM_solutions})}.\\
    }
    
    Construct lb-hypergraph $\mathbf{H}_{lb}$ by  \textbf{Equation (\ref{equation: elements_in_H_{lb}})}.\\
    
    Obtain complete hypergraph $\mathbf{H}$ by $\mathbf{H}=cat([\mathbf{H}_{saf},\mathbf{H}_{lb}],1)$.\\
    
    Embed Laplace based hypergraph manifold structure to dictionary learning objective function by \textbf{Equation (\ref{equation: laplace_regularization})}.\\
    
    \While{i < maxitem}
    {
        Update sparse coding $\mathbf{S}$ by \textbf{Equation (\ref{equation: optimization_UpdateS_Skn_1}),(\ref{equation: optimization_UpdateS_J})}.\\
        Update dictionary $\mathbf{D}$ by \textbf{Equation (\ref{equation: optimization_UpdateD})}.\\
    }
    
	\caption{Sparse Attention Hypergraph Regularized Dictionary Learning}
	\label{Algorithm: SAHDL}
\end{algorithm}



\subsection{Discussions}
In this subsection, we first explain the relation with existing hypergraph learning. Then we discuss the proposed SAHDL method in inductive and transductive learning.

\subsubsection{\textbf{Relation with HLSC}}
\label{section: Relation with HLSC}
Gao \emph{et al.} proposed HLSC \cite{gao2012laplacian} in 2012,  which is the first time to introduce the hypergraph Laplacian manifold to sparse coding. Compared with the HLSC method, the proposed SAHDL has mainly two contributions:

\textbf{$i)$} HLSC ignores the discriminative information, which is a vital guiding message for constructing atoms in a dictionary to represent original data. Specifically, the label guidance term, $\mathbf{H}_{lb}$, does not exist in HLSC. To this end, the SAHDL algorithm would be more suitable for supervised or semi-supervised learning task.

\textbf{$ii)$} In HLSC, all the samples play the same crucial roles when building hypergraph via feature similarity, e.g., the weight $z$ in Equation~(\ref{equation: elements_in_H_{saf}}) is equal to $1$. In contrast, the SAHDL algorithm offers attention weight for each sample to construct a more efficient connection. 

\textbf{$iii)$} In my opinion, compared with HLSC, the proposed SAHDL algorithm is more robust. As the reason that the sparse attention feature modal is helpful to mine the essential attributes, besides the guidance of label modal, SAHDL does not rely on the quality of features when constructing hypergraph. We evaluate this conclusion in section \ref{section: Robustness Analysis}.

\subsubsection{\textbf{SAHDL in Inductive and Transductive Learning}}
\label{section: inductive_transductive}
In pattern recognition tasks, the utilized data are usually split into the training set and the testing set. The approach mentioned above belongs to inductive learning since only employing training set feature embeddings for classification in the training stage. Different from inductive learning, transductive learning based methods use both training and testing set feature embeddings. To extend the SAHDL algorithm to transductive learning, we need to utilize all the feature embeddings when constructing the hypergraph and learning the dictionary. 

\section{Experiment}
In this section, to fairly evaluate the proposed SAHDL algorithm's effectiveness, we compared it with state-of-the-art dictionary learning methods on four benchmark datasets, including the UC Merced Land Use dataset, the RSSCN7 dataset, the Standford 40 Actions dataset, and the UIUC Sports Event dataset. We first describe the details of the adopted datasets. Then, we introduce the experimental setup and report the experimental results. Next, we conduct ablation studies to analyze the SAHDL method. Some discussions are listed in the end.

\subsection{Dataset}
The employed benchmark datasets can be split into two categories: 

\textbf{$i)$ Remote sensing datasets}, including UC Merced Land Use \cite{yang2010bag} and RSSCN7~\cite{zou2015deep}, the images in this kind of dataset exists a big difference among different classes. UC Merced Land Use (UCM-LU) dataset comes from the United States Geological Survey. It contains $2,100$ land-use images with $21$ categories, such as forest, intersection, medium-density residential, baseball diamond, etc. For the RSSCN7 dataset, it consists of $7$ classes of images, including parking lot, forest, lake region, etc. Each category is composed of $400$ images, and all of them come from Google Earth.

\textbf{$ii)$ Behavior identity datasets}, containing Stanford 40 Actions \cite{yao2011human} and UIUC Sports Event \cite{li2007and}, compared with remote sensing datasets, the gap among different classes of behavior identity dataset is not so big as that of the former. Stanford 40 Actions (Stanford40) dataset comprises $9,532$ human action images with $40$ categories, such as feeding a horse, riding a horse, riding a bike, repairing a bike, etc. Each category contains $180 \sim 300$ images, collected from Bing, Google, and Flickr. For UIUC Sports Event (UIUC-SE), there are $1,579$ sports event images with $8$ classes collected from the Internet, including snowboarding, sailing, rock climbing, etc. Each category consists of $137$ to $250$ diverse images.

We select $5$ samples from each category for training and $10$ samples for testing for all the datasets. More details of vertex and hyperedge are listed in the Table~\ref{table: dataset_split}.
\begin{table}
        \caption{Summary of datasets description}
        \label{table: dataset_split}
        \begin{center}
        \begin{tabular}{lcccc}
          \toprule
          \cmidrule(r){1-2}
          Dataset               & UCM-LU       & RSSCN7    & Stanford40    & UIUC-SE \\
          \midrule
          Vertices              & 2100         & 2800      & 9532          & 1579\\
          Training Vertices     & 105          & 35        & 200           & 40\\
          Testing Vertices      & 210          & 70        & 400           & 80\\
          Classes               & 21           & 7         & 40            & 8\\
          \bottomrule
        \end{tabular}
        \end{center}
\end{table}

\subsection{Experimental Setup}
For all the datasets, we employ standard Resnet~\cite{he2016deep} to extract feature embedding with $2,048$ dimensions. After that, we fix the dictionary size to $200$ and the nearest number of knn to $10$ for all datasets. The influence of dictionary size is discussed in the following section~\ref{section: ablation study}. In addition, there are three other parameters ($\epsilon$ in Equation~(\ref{equation: sparse_attention}), $\alpha$ and $\beta$ in Equation~(\ref{equation: Objective_function_inductive})) need to be tuned manually. Note that, the $\gamma$ in Equation~(\ref{equation: Classification}) is usually similar to the corresponding $\alpha$. 
Here, to evaluate the proposed SAHDL method fairly, we just show the optimal setups for best performance of SAHDL in inductive learning. Specifically, we set $\epsilon=2^{-6}$, $\alpha=2^{-6}$, $\beta=2^{3}$ for the UCM-LU dataset, $\epsilon=2^{-2}$, $\alpha=2^{-4}$, $\beta=2^{8}$ for the RSSCN7 dataset, $\epsilon=2^{-6}$, $\alpha=2^{-6}$, $\beta=2^{4}$ for the Stanford40 dataset, and $\epsilon=2^{-6}$, $\alpha=2^{-12}$, $\beta=2^{6}$ for the UIUC-SE dataset. For some discussions of parameters, we show them in the section~\ref{section: ablation study}. 

\subsection{Experimental Results}
\begin{table*}
        \caption{Classification results}
        \label{table: classification_results}
        \begin{center}
            \begin{tabular}{lcccc}
                \toprule
                Methods$\backslash$Datasets                     & UCM-LU        & RSSCN7        & Stanford40    & UIUC-SE   \\
                \midrule
                SRC (TPAMI \cite{wright2009robust}, 2009)       & 80.4$\%$      & 67.1$\%$      & 60.5$\%$      & 92.7$\%$  \\
                CRC (ICCV \cite{zhang2011sparse}, 2011)         & 80.7$\%$      & 67.7$\%$      & 58.8$\%$      & 92.5$\%$  \\
                NRC (PR \cite{xu2019sparse}, 2019)              & 81.6$\%$      & 69.7$\%$      & 60.1$\%$      & 91.7$\%$  \\
                SLRC (TPAMI \cite{deng2018face}, 2018)          & 81.0$\%$      & 66.4$\%$      & 60.4$\%$      & 90.8$\%$  \\
                Euler-SRC (AAAI \cite{liu2018euler}, 2018)      & 80.9$\%$      & 69.7$\%$      & -             & -  \\
                LC-KSVD (TPAMI \cite{jiang2013label}, 2013)     & 79.4$\%$      & 68.0$\%$      & -             & -  \\
                CSDL (NC \cite{liu2016face}, 2016)              & 80.5$\%$      & 66.7$\%$      & 57.6$\%$      & 93.4$\%$  \\
                LC-PDL (IJCAI \cite{zhang2019scalable}, 2019)   & 81.2$\%$      & 69.7$\%$      & 61.2$\%$      & 95.1$\%$  \\
                FDDL (ICCV \cite{yang2011fisher}, 2011)         & 81.0$\%$      & 64.0$\%$      & -             & - \\
                LEDL (NC \cite{shao2020label}, 2020)            & 80.7$\%$      & 67.9$\%$      & 60.7$\%$      & 92.5$\%$  \\
                ADDL (TNNLS \cite{zhang2018jointly}, 2018)      & 83.2$\%$      & 72.3$\%$      & 60.9$\%$      & 94.3$\%$  \\
                CDLF (SP \cite{wang2020class}, 2020)            & 81.0$\%$      & 69.6$\%$      & 60.0$\%$      & 94.9$\%$  \\
                HLSC (TPAMI \cite{gao2012laplacian}, 2012)      & 81.4$\%$      & 71.1$\%$      & 60.6$\%$      & 94.3$\%$  \\
                \midrule
                SAHDL                                         & \bf{85.2$\%$} & \bf{75.7$\%$} & \bf{61.5$\%$} & \bf{96.3$\%$} \\
                \bottomrule
            \end{tabular}        
        \end{center}
\end{table*}
We compare the SAHDL approach with several classical classification methods, including SRC \cite{wright2009robust}, CRC \cite{zhang2011sparse}, NRC \cite{xu2019sparse}, SLRC \cite{deng2018face}, Euler-SRC \cite{liu2018euler}, LC-KSVD \cite{jiang2013label}, CSDL \cite{liu2016face}, LC-PDL \cite{zhang2019scalable}, FDDL \cite{yang2011fisher}, LEDL \cite{shao2020label}, ADDL \cite{zhang2018jointly}, CDLF \cite{wang2020class}, HLSC \cite{gao2012laplacian}. We show the experimental results in Table~\ref{table: classification_results}, and obtain the following observations.

\textbf{$i)$} 
The proposed SAHDL approach obtains outstanding classification performances on all four datasets. On the two remote sensing datasets, UCM-LU and RSSCN7, the classification rates of the SAHDL approach outperform other methods at least $2\%$ and $3.4\%$, respectively. And on the two behavior identity datasets, the classification rates of the SAHDL approach achieve the best performance with an improvement of at least $0.3\%$ on the Stanford40 dataset and $1.4\%$ on the UIUC-SE dataset.

\textbf{$ii)$} 
Compared with several traditional classification methods (e.g., without dictionary learning module), even the latest approaches in recent years, such as NRC, SLRC, Euler-SRC, the classification rates of the proposed SAHDL approach exceed them at least $3.6\%$ on two remote sensing datasets and $1.0\%$ on two behavior identity datasets. However, there also exists a disadvantage. The SAHDL approach consumes more time than these traditional methods. For example, the SAHDL approach expends about $35$ seconds for a complete process on the UCM-LU dataset, while NRC, SLRC, Euler-SRC cost less than $10$ seconds.

\textbf{$iii)$} 
The SAHDL approach also achieves the highest accuracy among dictionary learning based works, including LC-KSVD, CSDL, LC-PDL, FDDL, LEDL, ADDL, CDLF, HLSC. Still, sometimes it does not improve much, especially on the Stanford40 dataset. The SAHDL approach achieves similar performance to the LC-PDL method. The reason is that all the dictionary learning based methods listed in Table~\ref{table: classification_results} have their highlights. The SAHDL approach only introduces the sparse attention hypergraph manifold structure to a basic dictionary learning model. In other words, the proposed SAHDL approach is a model-agnostic module, which can be embedded in any dictionary learning based works to promote the performance.

\textbf{$iv)$} 
The SAHDL approach can be regarded as the improvement of HLSC, while from the Table~\ref{table: classification_results}, we can see that the classification rate of the SAHDL approach outperforms that of the HLSC method at least $0.9\%$ on all four datasets. It has demonstrated the effectiveness of the proposed SAHDL method. Please refer to section \ref{section: ablation study} to concern the reasons for the improvement.

\subsection{Ablation Study}
\label{section: ablation study}
The SAHDL approach has achieved outstanding performance. It is interesting to recognize what are the factors affecting the experimental results. For this purpose, we design three ablation studies to discuss the proposed SAHDL method.
\begin{figure}
	\begin{center}
		\includegraphics[width=1.0\linewidth]{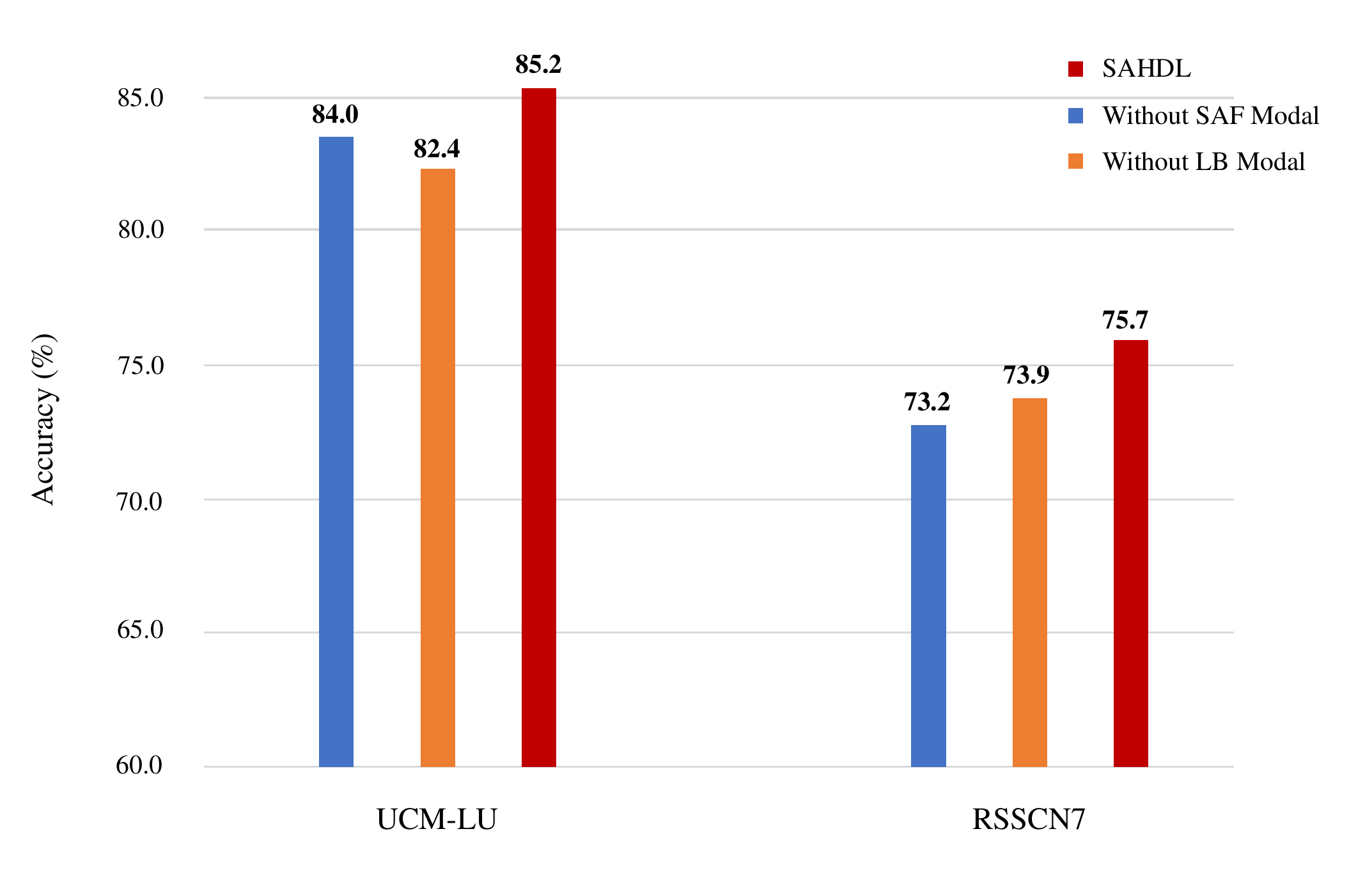}
	\end{center}
	\caption{The influence of saf-modal and lb-modal.}
	\label{figure: saf-lb}
\end{figure}

\textbf{$i)$}
As mentioned above, compared with the HLSC method, we have two contributions to promote performance. Here, we first evaluate the influence of sparse attention feature modal on two remote sensing datasets, e.g., we set the $z_i$ in Equation~\ref{equation: elements_in_H_{saf}} to $1$. After that, we try to remove the label modal to demonstrate the importance of the guidance of discriminative information, e.g., we set $\mathbf{H}=\mathbf{H}_{saf}$. Figure~\ref{figure: saf-lb} shows the comparison. It is clear that no matter which part is absent, it would greatly impact the accuracy. Notably, the SAHDL approach without the SAF modal or the LB modal is still superior to the HLSC method.
\begin{figure}
	\begin{center}
		\includegraphics[width=1.0\linewidth]{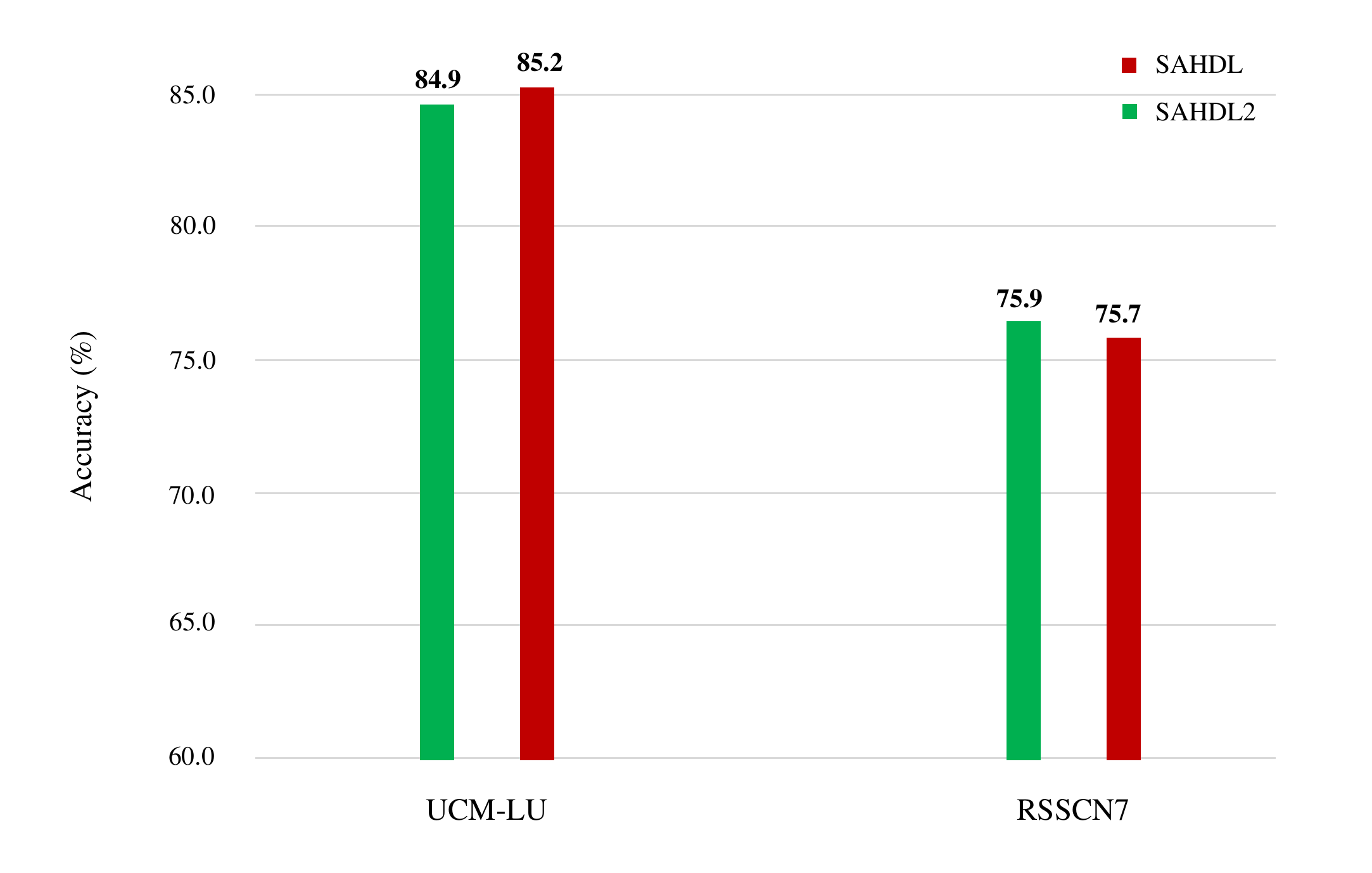}
	\end{center}
	\caption{Comparison with SAHDL2.}
	\label{figure: SAHDL2}
\end{figure}
\begin{figure*}
    \subfigure[]{
        \begin{minipage}[t]{0.3\linewidth}
        	\begin{center}
        		\includegraphics[width=0.8\linewidth]{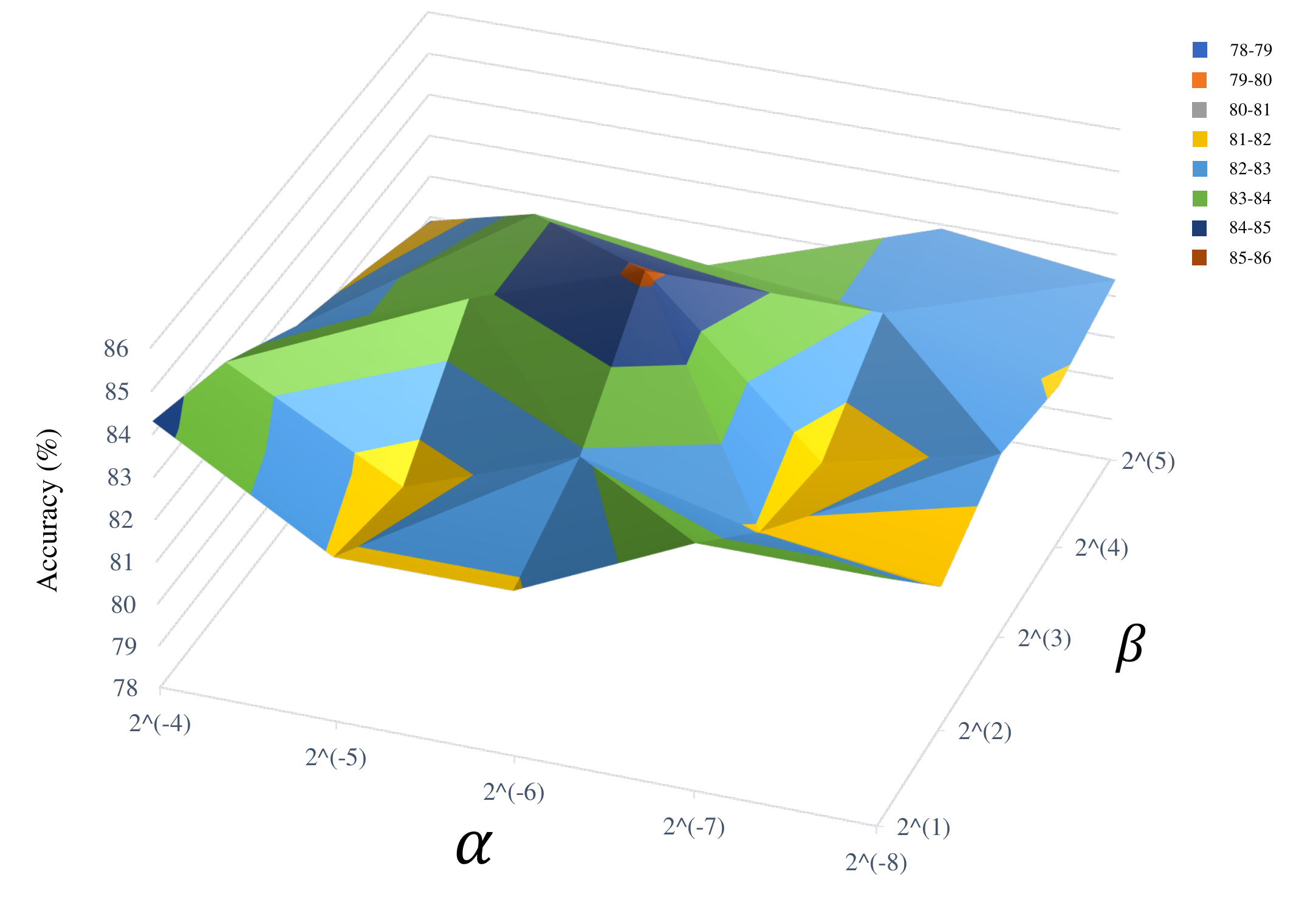}
        	\end{center}
        	\label{figure: alpha_beta}
        \end{minipage}%
    }%
    \subfigure[]{
        \begin{minipage}[t]{0.3\linewidth}
        	\begin{center}
        		\includegraphics[width=0.8\linewidth]{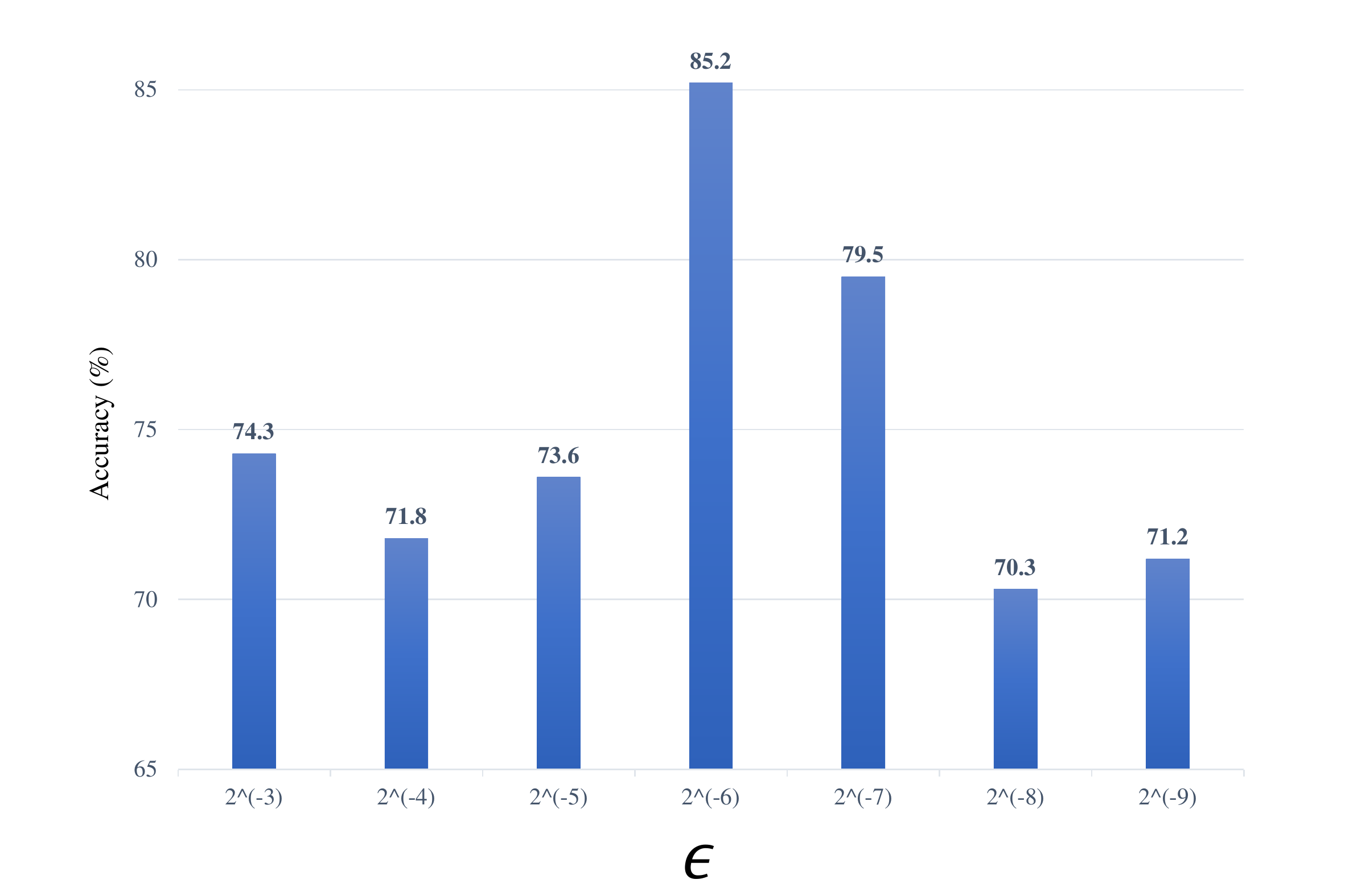}
        	\end{center}
        	\label{figure: epsilon}
        \end{minipage}%
    }%
    \subfigure[]{
    \begin{minipage}[t]{0.3\linewidth}
    	\begin{center}
    		\includegraphics[width=0.8\linewidth]{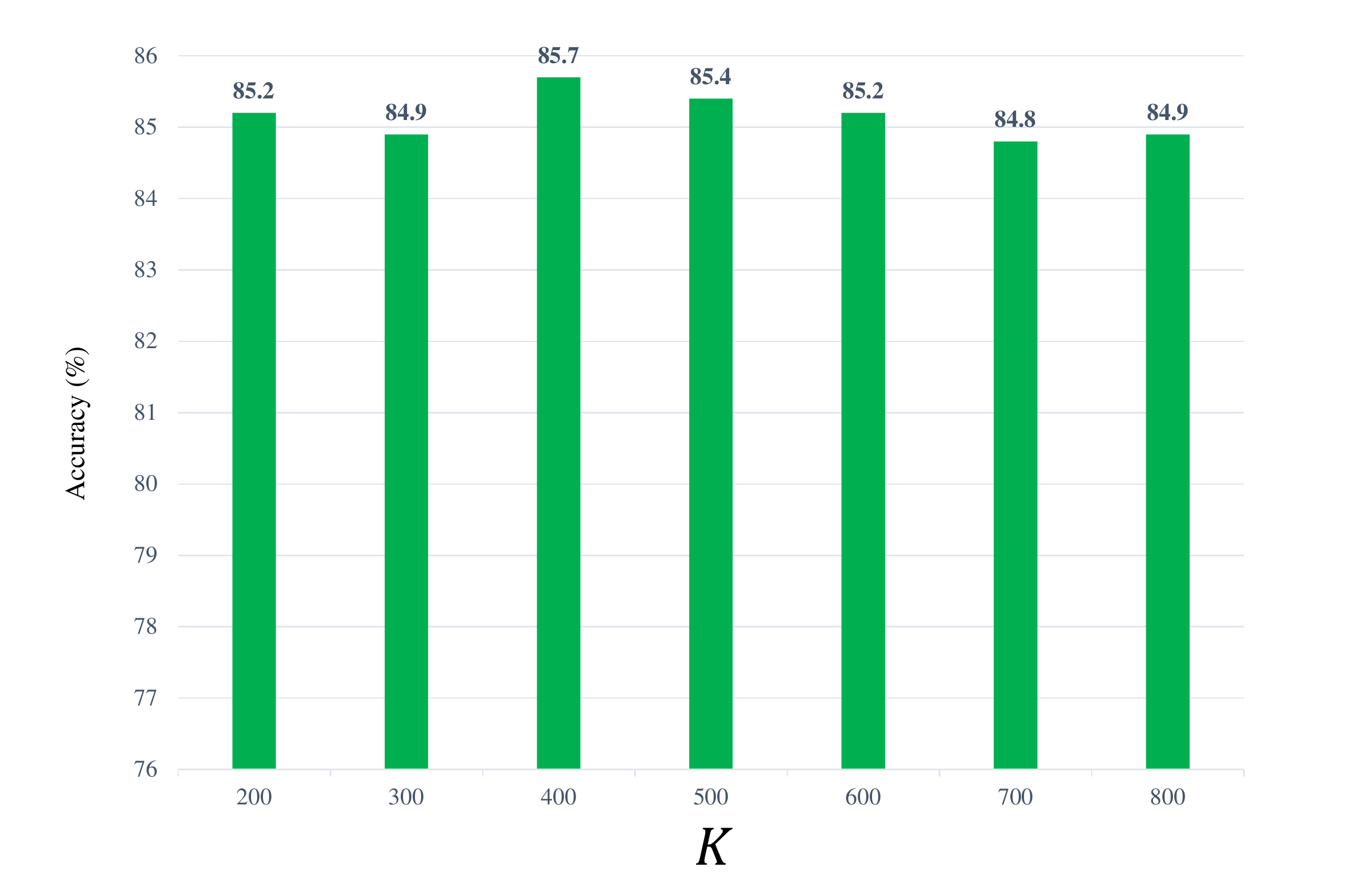}
    	\end{center}
    	\label{figure: K}
    \end{minipage}%
    }%
    \caption{The influence of parameters}
    \label{figure: alpha_beta_epsion_K}
\end{figure*}

\textbf{$ii)$}
In section~\ref{Section: Introduction}, several ways of introducing discriminative information to the objective function have been discussed. Besides the proposed SAHDL approach, there are other classical methods, e.g., directly adding the label constraint term into the objective function. We name this method as SAHDL2 and summarize it as follows:
\begin{equation}
\begin{split}
        &\mathop {\arg \min}\limits_{\mathbf{D},\mathbf{S},\mathbf{B}}     
        \mathcal{F}(\mathbf{D},\mathbf{S},\mathbf{B})\\
		&=\left\| \mathbf{X} - \mathbf{D}\mathbf{S} \right\|_F^2 
		+ 2\alpha \left\| \mathbf{S} \right\|_{\ell_1} 
		+ \beta {\kern 2pt} \text{tr} \left( \mathbf{\Delta}_l \mathbf{S}^T \mathbf{S} \right)
		+ \omega \left\| \mathbf{U} - \mathbf{B}\mathbf{S_{l}} \right\|_F^2\\
		&{\rm{s}}.t.\left\| {{{\mathbf{d}}_{k}}} \right\|_2^2 \le 1, {\kern 4pt} \left\| {{{\mathbf{b}}_{k}}} \right\|_2^2 \le 1 {\kern 4pt} \left( {k = 1,2, \cdots K} \right)\\
\end{split}
\label{equation: Objective_function_label}
\end{equation}
where $\mathbf{B} \in \mathbb{R}^{C \times K}$ is the classifier plane, $C$ is the class number. $\mathbf{U} \in \mathbb{R}^{C \times N_{l}}$ denotes the label matrix, $N_l$ represents the number of labeled training data. $\mathbf{S}_l \in \mathbb{R}^{K \times N_{l}}$ denotes the labeled parts of $\mathbf{S}$. In addition, $\mathbf{\Delta}_l=\mathbf{I} - \mathbf{D}_v^{-\frac{1}{2}} \mathbf{H}_{saf} \mathbf{W} \mathbf{D}_e^{-1} {\mathbf{H}_{saf}}^T \mathbf{D}_v^{-\frac{1}{2}}$.  

To further evaluate the effectiveness of discriminative information in the SAHDL approach, we compare it with the SAHDL2 (e.g., Equation~(\ref{equation: Objective_function_label})) on two remote sensing datasets. Figure~\ref{figure: SAHDL2} shows the comparative experimental results. The SAHDL approach achieves similar performance compared with the SAHDL2. Notably, in Equation~(\ref{equation: Objective_function_label}), we can adjust the parameter $\omega$ to control the label constraint term. However, in the proposed SAHDL approach, we fix all the hyperedge weights to $1$. That is to say, we can obtain better performance if we adjust the $\mathbf{W}$ (described in section \ref{Section: Hypergraph_Construction}) for each hyperedge. 

\textbf{$iii)$}
For the proposed SAHDL approach, three parameters (e.g., $\epsilon$, $\alpha$, $\beta$) mainly affect the experimental results. For the dictionary size, we set it to $200$ for all the datasets. We can achieve similar results with different reasonable dictionary size according to adjusting other parameters. To further evaluate these conclusions, we conduct three comparative experiments on the UCM-LU dataset. 

More specifically, $\epsilon$ influences the performance independently. Thus we fix $\alpha$ and $\beta$ to observe the effect of $\epsilon$, and Figure~\ref{figure: alpha_beta_epsion_K} shows the experimental results. Different $\epsilon$ corresponds to different $\mathbf{z}$. We can observe that the experimental results are sensitive to this parameter. Only the optimal $\epsilon$ (i.e., only the optimal $\mathbf{z}$) is helpful for the performance. This phenomenon also proves the importance of attention weights.

For $\alpha$ and $\beta$, they interact with each other. Thus we fix $\epsilon$, and explore the impact of these two parameters simultaneously. Figure~\ref{figure: alpha_beta_epsion_K} shows the experimental results. The proposed SAHDL approach has strong adaptability to these two parameters. We can flexibly choose the pairwise $\alpha$ and $\beta$ to obtain the best performance. Following, we evaluate the influence of dictionary size $K$. In this experiment, we tune the parameters $\epsilon$, $\alpha$, and $\beta$ to obtain the optimal performance for each dictionary size. Figure~\ref{figure: alpha_beta_epsion_K} shows the experimental results. From Figure~\ref{figure: alpha_beta_epsion_K}, we can conclude that the proposed SAHDL approach is not sensitive to the dictionary size. Thus, we would choose a small size to reduce training time.
\begin{figure}
	\begin{center}
		\includegraphics[width=1.0\linewidth]{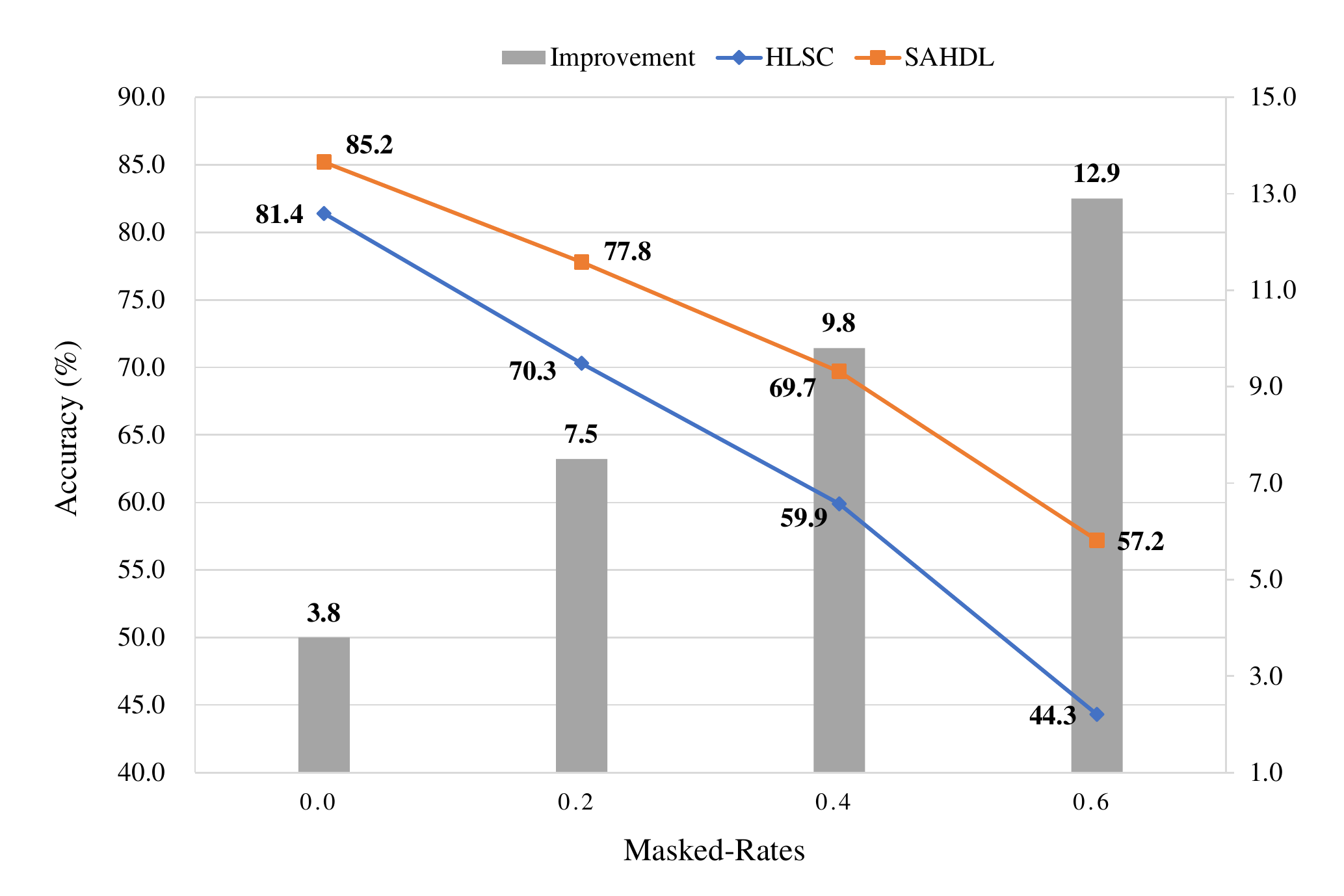}
	\end{center}
	\caption{The influence of adding mask to SAHDL.}
	\label{figure: mask}
\end{figure}
\subsection{Robustness Analysis}
\label{section: Robustness Analysis}
As mentioned in section~\ref{section: Relation with HLSC}, the proposed SAHDL approach is more robust than the HLSC method. To demonstrate this conclusion, we randomly generate multiple masks to add noises for each sample. Figure~\ref{figure: mask} shows the verification result. Different values on the abscissa denote the extent of masked-feature. From the figure, we can see that the improvement of classification results changed with different masked-feature. As masked-feature increases ($0.0$, $0.2$, $0.4$, $0.6$), the noises in feature embedding increase, and the accuracy's improvement is more significant for the proposed SAHDL approach ($3.5\%$, $7.5\%$, $9.8\%$, $12.9\%$). It indicates SAHDL is more suitable for the datasets without high-quality features.

\subsection{Discussion}
Compared with several state-of-the-art methods, the experimental results have demonstrated that the proposed sparse attention hypergraph plays a positive role in dictionary learning. Also, we conclude that the SAHDL approach is sensitive to the attention weights but insensitive to the features' quality. Besides, the proposed sparse attention mechanism is a model-agnostic method. We can flexibly expand it to any standard models. 

\section{Conclusion}
The attention mechanism on a deep network usually updates attention weights with the network propagating. Few attempts aim to introduce the attention mechanism to traditional machine learning. We propose a novel sparse attention mechanism for hypergraph and introduce it into the dictionary learning procedure to tackle this issue. We also incorporate the label information into the hypergraph. The outstanding performance on four benchmark datasets has demonstrated the efficiency of the proposed SAHDL approach. It would be interesting for future work to expand the sparse attention mechanism to a deep neural network.

\vfill\pagebreak

\bibliographystyle{IEEEbib.bst}
\bibliography{refs.bib}

\end{document}